%% file: main.tex
\documentclass{bmvc2k}

\input{plots}

\usepackage{graphicx}
\usepackage{amsmath}
\usepackage{amssymb}
\usepackage{booktabs}
\usepackage[subtle]{savetrees}  
\usepackage{float}

\usepackage[capitalize]{cleveref}
\crefname{section}{Sec.}{Secs.}
\Crefname{section}{Section}{Sections}
\Crefname{table}{Table}{Tables}
\crefname{table}{Tab.}{Tabs.}

\input{preamble_bmvc}
\begin{document}

\input{sec/0_metadata}
\maketitle
\input{sec/0_abstract_bmvc}
\input{sec/1_introduction_bmvc}

\input{sec/2_related_bmvc}
\input{sec/4_method_bmvc}
\input{sec/5_results_bmvc}

\input{sec/6_conclusions_bmvc}

{
    \bibliography{egbib}
}


\end{document}

%% file: plots.tex
\usepackage{tikz}
\usetikzlibrary{arrows,shapes,calc,matrix,fit,backgrounds}
\usepackage{pgfplots}
\usepackage{pgfplotstable}
\pgfplotsset{compat=1.9}

\usepackage{xstring}

\usepgfplotslibrary{external}

\newcommand{\extdata}[1]{\input{#1}}
\IfBeginWith*{\jobname}{fig/extern/}{\finalcopy}{}

\newcommand{\leg}[1]{\addlegendentry{#1}}

\tikzset{every mark/.append style={solid}}
\pgfplotsset{
	grid=both, width=\columnwidth, try min ticks=5,
	every axis/.append style={font=\scriptsize},
	every axis plot/.append style={thick,mark=none,mark size=1.2,tension=0.18},
	legend cell align=left, legend style={fill opacity=0.8},
}

\pgfplotsset{
	dash/.style={mark=o,dashed,opacity=0.7},
	dott/.style={mark=o,dotted,opacity=0.7},
}

%% file: preamble_bmvc.tex
\newcommand{\acro}{\mathrm{MTN}}
\newcommand{\MTN}{$\acro$\xspace}
\newcommand{\MTNcls}{$\acro_{\text{CLS}}$\xspace}
\newcommand{\MTNnn}{$\acro_{\text{NN}}$\xspace}
\newcommand{\MTNmix}{$\acro_{\text{MIX}}$\xspace}

\usepackage{overpic}
\usepackage{enumitem} 
\usepackage{overpic} 
\usepackage{color}
\usepackage{fontawesome}

\definecolor{turquoise}{cmyk}{0.65,0,0.1,0.3}
\definecolor{purple}{rgb}{0.65,0,0.65}
\definecolor{dark_green}{rgb}{0, 0.5, 0}
\definecolor{orange}{rgb}{0.8, 0.6, 0.2}
\definecolor{red}{rgb}{0.8, 0.2, 0.2}
\definecolor{darkred}{rgb}{0.6, 0.1, 0.05}
\definecolor{blueish}{rgb}{0.0, 0.3, .6}
\definecolor{light_gray}{rgb}{0.7, 0.7, .7}
\definecolor{pink}{rgb}{1, 0, 1}
\definecolor{greyblue}{rgb}{0.25, 0.25, 1}

\newcommand{\loss}{\mathcal{L}}

\newcommand{\D}{\mathcal{D}}
\newcommand{\M}{\mathcal{M}}
\newcommand{\kl}[2]{\mathcal{D}_{KL}(#1 ~||~ #2)}

\newcommand{\nn}[1]{\ensuremath{\text{NN}_{#1}}\xspace}

\def\l1{\ensuremath{\ell_1}\xspace}
\def\l2{\ensuremath{\ell_2}\xspace}

\newcommand{\commentout}[1]{}

\newcommand{\head}[1]{{\noindent\bf #1}}

\newcommand{\real}{\mathbb{R}}

\newcommand{\defn}{\mathrel{:=}}

\def\msp{\hspace{5pt}}
\def\lsp{\hspace{12pt}}

\newcommand{\cC}{\mathcal{C}}
\newcommand{\cD}{\mathcal{D}}

\newcommand{\cM}{\mathcal{M}}

\newcommand{\cX}{\mathcal{X}}
\newcommand{\cY}{\mathcal{Y}}

\newcommand{\vp}{\mathbf{p}}
\newcommand{\vq}{\mathbf{q}}

\newcommand{\vy}{\mathbf{y}}
\newcommand{\vz}{\mathbf{z}}

\def\onedot{.\xspace}
\def\eg{\emph{e.g}\onedot} 
\def\ie{\emph{i.e}\onedot}

\def\etal{\emph{et al}.}

\definecolor{F7E0D5}{RGB}{247,224,213}
\colorlet{Light}{white!0!F7E0D5}
\makeatother

%% file: sec/0_metadata.tex
\title{A Memory Transformer Network for Incremental Learning}

\addauthor{Ahmet Iscen}{}{1}
\addauthor{Thomas Bird}{}{2}
\addauthor{Mathilde Caron}{}{1}
\addauthor{Alireza Fathi}{}{1}
\addauthor{Cordelia Schmid}{}{1}

\addinstitution{
Google Research
}
\addinstitution{
University College London
}

\runninghead{Iscen \etal}{MTN for Incremental Learning}

%% file: sec/0_abstract_bmvc.tex
\begin{abstract}

We study class-incremental learning, a training setup in which new classes of data are observed over time for the model to learn from. 
Despite the straightforward problem formulation, the naive application of classification models to class-incremental learning results in the ``catastrophic forgetting" of previously seen classes. 
One of the most successful existing methods has been the use of a memory of exemplars, which overcomes the issue of catastrophic forgetting by saving a subset of past data into a memory bank and utilizing it to prevent forgetting when training future tasks. 
In our paper, we propose to enhance the utilization of this memory bank: we not only use it as a source of additional training data like existing works but also integrate it in the prediction process explicitly.
Our method, the Memory Transformer Network (MTN), learns how to combine and aggregate the information from the nearest neighbors in the memory with a transformer to make more accurate predictions.
We conduct extensive experiments and ablations to evaluate our approach.
We show that MTN achieves state-of-the-art performance on the challenging ImageNet-1k and Google-Landmarks-1k incremental learning benchmarks.

\end{abstract}

%% file: sec/1_introduction_bmvc.tex
\section{Introduction}
\label{sec:intro}
\input{fig/teaser_bmvc}

The goal of Incremental Learning (IL) is to design models capable of dynamically learning when the input data becomes available gradually over time.
The main challenge is to adapt to the new incoming data, while still being able to remember the previously learned knowledge.
In the past, many IL approaches have been proposed to avoid \emph{catastrophic forgetting}, which refers to the situation where an incremental learner ends up performing poorly on previously learned tasks \cite{goodfellow2014iclr}. One popular and successful way of approaching catastrophic forgetting is to keep a small subset of training data from earlier tasks, \ie \textit{exemplars} stored in a \textit{memory bank}, and use them to augment the data when training on the new task, which is referred as \textit{rehearsal} \cite{robins1995rehearsal}.
This approach has been shown to be very effective for IL and, consequently, the use of a memory to learn the model parameters constitutes a key component of most modern methods~\cite{rebuffi2017icarl,hou2019rebalance,belouadah2019il2m,wu2019large,iscen2020memoryefficient,ahn2020ssil}.

In our work, we propose to go one step further and exploit the exemplars not only as a source of additional training data but also in the prediction process itself.
Our hypothesis is that the relationships between a query (\ie. a training or testing example) and the different elements of the memory bank generate a strong signal, which helps the incremental learner make a more robust and accurate prediction with respect to the given query.
To this end, we propose the \emph{Memory Transformer Network} (\MTN) which looks at the local feature neighborhood of a query vector before making a prediction.
\MTN is a light-weight transformer that makes a class prediction for a given query by directly modeling the relationship between this query and the feature representations of the exemplars in a memory bank.
Since conditioning on the entire memory bank would be too computationally demanding, we choose to feed \MTN only a reduced set of exemplars (selected with nearest neighbour search).

Our approach allows rare patterns to be memorized explicitly, rather than implicitly in model parameters.
The prediction of a query does not only depend on its representation anymore, it also depends on its nearest neighbors amongst the preserved exemplars.
This makes the model more accurate by directly retrieving the corner cases from the memory, instead of dedicating the model parameters to remember them.

\MTN is inspired from recent \textit{memory transformer} architectures in language modeling~\cite{wu2022memorizing} and video recognition~\cite{wu2022memvit}, where the input sequence to the transformer follows a natural ordering, \eg sequence of words and frames. 
In these frameworks, the memory typically contains previously seen words and frames which helps the system make predictions.
\MTN extends memory transformers to image classification, where we define the input sequence based on the local neighborhood of the query in the feature space.

To summarize, our contributions are as follows:
\begin{itemize}
    \item We believe our approach is the first to propose a model with external memory for class-incremental learning. We do so by utilizing a transformer that combines the query input features with the memory exemplar features.
    \item Unlike the existing works on class-incremental learning, we leverage the exemplars not only during the model training, but also directly in the process of decision making by having the output distribution depend on the external memory.
    \item Our method achieves state-of-the-art results on ImageNet-1k~\cite{DSLLF09} and Google-Landmarks-1k~\cite{weyand2020google} datasets.
\end{itemize}

%% file: fig/teaser_bmvc.tex
\begin{figure}
\begin{center}
\includegraphics[width=0.9\linewidth]{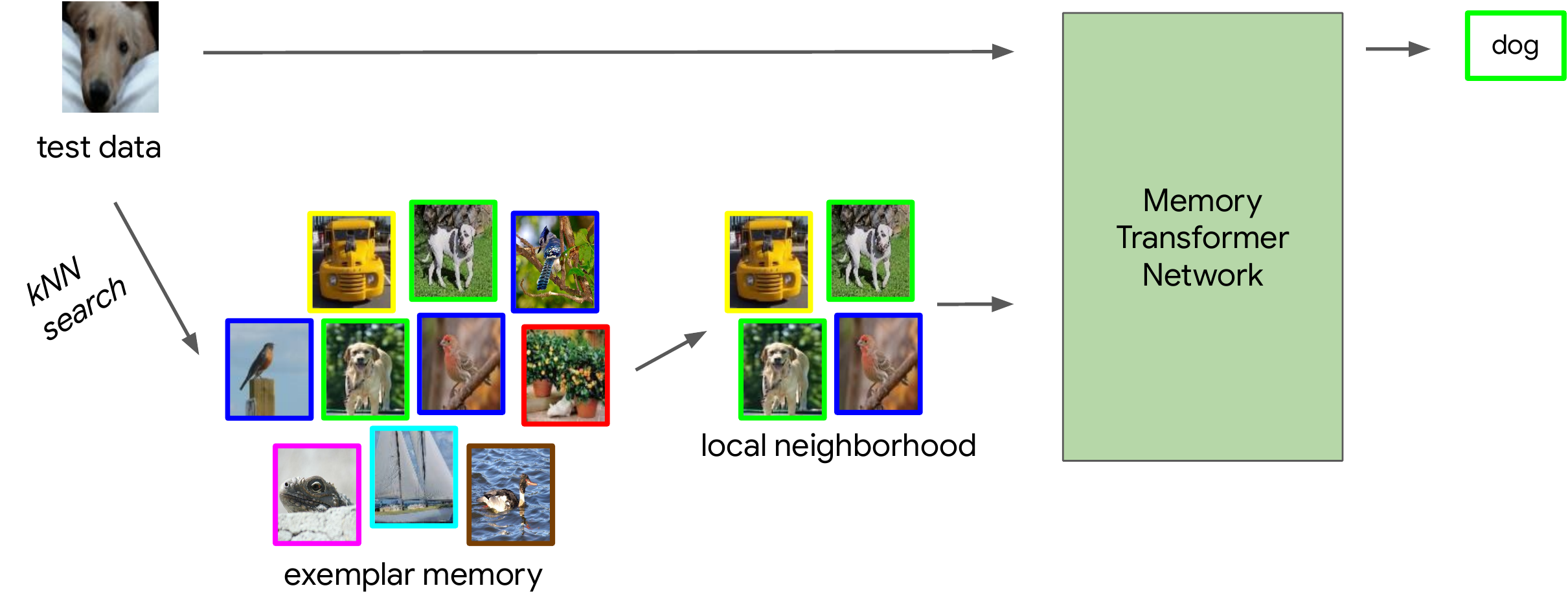}
\end{center}
\caption{
%
The Memory Transformer Network (\MTN) uses a memory of data from previous and current tasks to help classify incoming data. 
Feature representations of images are stored in the exemplar memory.
During the inference, kNN search is performed to find the local neighborhood of the test query vector.
\MTN then aggregates the information from the local neighborhood to make its prediction.
}
\label{fig:teaser}
\end{figure}

%% file: sec/2_related_bmvc.tex
\section{Related works}
\label{sec:related}

In this section we discuss previous works on class-incremental learning, looking at both exemplar-based and exemplar-free methods. We also discuss methods in the literature which augment neural networks with memory, which is at the heart of our method.

A popular method in class-incremental learning is to keep a subset of a task training data as exemplars, and use it in subsequent tasks to mitigate the forgetting effect as incremental training proceeds. Adding the exemplars to the current task training set is one effective way to prevent forgetting, and is referred to as rehearsal \cite{robins1995rehearsal}. Rebuffi~\etal~propose the iCaRL method \cite{rebuffi2017icarl}, which selects the exemplars via a simple algorithm known as herding and uses the mean exemplar vector for each class at inference time rather than the classifier used during training. Recent works extend iCaRL, and have focused on correcting the model bias towards tasks later on in incremental training. Some methods train additional parameters \cite{wu2019large,yan2021dynamically}, while others amend the loss function \cite{hou2019rebalance,belouadah2019il2m,tao2020tpcil,ahn2020ssil, mittal2021essentials}. There have also been explorations into optimizing the exemplars themselves \cite{liu2020mnemonics}, and architectural changes like splitting the model into stable and plastic weights \cite{Liu2020AANets} to allow for fast task training but stable knowledge of past tasks. To reduce the memory footprint of saving exemplars as images, Iscen~\etal~\cite{iscen2020memoryefficient} store features rather than images, adapting old exemplar features to be compatible with new features via a separate network.

There are also many works in class-incremental learning which do not utilize exemplars. Li and Hoiem \cite{li2017learning} add a knowledge distillation loss to the usual classification loss, helping to prevent forgetting as tasks are seen. Lopez-Paz and Ranzato \cite{lopz2017episodic} modify the gradient update at a given task, and also pass updates to previously seen classes. There are also a variety of methods which learn a generative model to simulate data from past tasks, a method known as \textit{pseudo-rehearsal}. Multiple methods explore using adversarial training to generate the pseudo-exemplars \cite{shin2017continual,wu2019memory}, \cite{xiang2019adversarial}, whereas Smith~\etal~\cite{smith2021dreaming} use model-inversion to generate the fake data.

Santoro~\etal~\cite{santoro2016mann} describe a memory-augmented neural network (MANN), in which a differentiable external memory is learned in order to improve neural network performance on meta-learning. There are a number of works in the NLP domain that allow the model to utilize an external memory bank of knowledge \cite{fevry2020entities,guu2020realm}. Liu~\etal~\cite{liu2018unsupervised} use a memory implemented through a Hopfield network and pretrained features on unsupervised learning. Mi and Faltings \cite{mi2020memory} use a MANN in incremental learning, on the application of recommendation systems. There are a number of works which resemble learned versions of nearest neighbors, predicting labels by calculating similarities of a query point to a support set of features and labels \cite{vinyals2016matching,snell17prototypical,ploetz2018nnn}. There are also recent works that use transformers to process the query points relation to memory. Gordo~\etal~\cite{gordo2020attentionbased} apply a transformer for retrieval, adapting a query point and its nearest neighbors. Doersch et al. \cite{doersch2021crosstransformers} use a transformer in meta-learning, predicting a label for a query based on the spatially-aware similarities to labelled points from memory. 

%% file: sec/4_method_bmvc.tex
\section{Method}
\input{fig/overview}

In this section, we begin by describing the problem definition and the notation used throughout this paper. We then introduce our method in Section~\ref{sec:tcil}.

\subsection{Background and notation}
\label{sec:background}

We define the dataset $\cD = \{(x,y)|x \in \cX, y\in\cY \}$ which consists of a set of images $\cX$ and their corresponding labels $\cY$.
We assume that the labels $\cY$ belong to one of the classes in $\cC$, which is a set of $C$ classes.

In incremental learning, we do not have access to the entire dataset at once.
Instead, the dataset is split into $T$ tasks, where each task contains disjoint classes, \ie $\cC^1, \cC^2, \dots, \cC^T$, where $\cC = \cC^1 \cup \cC^2 \cup \dots \cup \cC^T$ and $\cC^i \cap \cC^j = \emptyset$ for $i \neq j$. 
At task $t$, we only have access to the data corresponding to the set of classes $\cC^t$, and the training data in task $t$ is defined by $\cD^t = \{(x,y)|x \in \cX^t, y\in\cY^t \}$.

The learned model typically consists of two parts: a \emph{feature extractor}  and a \emph{classifier}.
The feature extractor $f$ maps each input image $x$ to a $d$-dimensional feature vector, \ie $\vq \defn f(x) \in \real^d$.
Consequently, the \emph{classifier} $g: \real^d \rightarrow \real^C$ is applied to each vector $\vq$ to obtain class prediction scores, or \emph{logits}, denoted by $\vz$, \ie $\vz \defn g(\vq) \in \real^C$. We denote the model parameters used in the feature extractor and classifier as $\theta$.

The best-performing recent methods in class-incremental learning have been rehearsal-based methods \cite{rebuffi2017icarl,hou2019rebalance,belouadah2019il2m,wu2019large,iscen2020memoryefficient,ahn2020ssil}.
That is, they rely on saving some subset of class training data $\cM^{t} \subset \cD^t$, \ie the \emph{exemplars}.
Exemplars can be stored as images~\cite{rebuffi2017icarl,hou2019rebalance,belouadah2019il2m,wu2019large,ahn2020ssil}, or features~\cite{iscen2020memoryefficient}.
The training data in task $t$ contains all the exemplars from the previous tasks in order to prevent catastrophic forgetting, \ie $\D^t_{\text{train}} = \D^t \cup \M^{1:t-1}$.
Typically, the number of exemplars kept is significantly less than the total number of samples, \ie $|\M^t| <<|\D^t|$.

Recent works show that alternatives to the usual cross-entropy loss can benefit training in class-incremental learning \cite{hou2019rebalance,belouadah2019il2m,ahn2020ssil}. In this work, we use the separated-softmax loss \cite{ahn2020ssil}, which splits the classification loss into two terms:
\begin{align}\label{eqn:ssloss}
    \loss_{SS}(x, y) &=~ \delta_{y \in \cC^t} \kl{\vy^t}{\vp^t} ~+ \nonumber\\
    &\delta_{y \in \cC^{1:t-1}}  \kl{\vy^{1:t-1}}{\vp^{1:t-1}},
\end{align}
where $\vy^t$ is the one-hot vector giving the label $y$ in the task classes $\cC^t$, and 
$\vp^t = \text{softmax}(\vz^t)$ is the vector of probabilities, over the same classes, produced by the model.
We also augment the classification loss(Eq.~\eqref{eqn:ssloss}) with a task-wise distillation loss, which has been shown to be effective in preventing forgetting during class-incremental learning \cite{li2017learning,castro2018endtoend}:
\begin{equation}\label{eq:tkd}
    \loss_{TKD}(x) = \sum_{i=1}^{t-1}\kl{\vp^{i}_{t-1}}{\vp^{i}},
\end{equation}
where ${\vp^{i}_{t-1}}$ is obtained with the frozen model parameters  $\theta_{t-1}$ from the end of the previous task training. Overall our objective is:
\begin{equation}
    \loss(x, y) = \loss_{SS}(x, y) + \loss_{TKD}(x).
\end{equation}

\subsection{Memory Transformer Network}
\label{sec:tcil}
Existing works only use exemplars to prevent catastrophic forgetting when training the model classifier $g$. Our work is based on the observation that, since we have to maintain the memory of exemplars, we can also utilize the exemplar memory to improve our predictions.

We define the memory by matrix $V$, an $M \times d$ matrix containing the feature vectors, extracted for each exemplar with a feature extractor $f$. We drop the task indices $t$ here for simplicity.
Typically, the memory $V$ is used to rehearse the information from the previous tasks while learning the model parameters $\theta$ through the training process in incremental learning.
The memory has no impact during the actual prediction, which follows the distribution 
$p_{\theta}(y|\vq)$, where $y$ is the class label given the corresponding query vector $\vq$.

However, given the catastrophic forgetting effect exhibited by neural networks when learning incrementally, we instead use $V$ explicitly in the prediction process. That is, we model the output label distribution as $p_{\theta}(y|\vq, V)$, which now depends on the memory $V$ as well as the query $x$.
To incorporate the exemplar memory $V$ into the prediction process, we add a layer between feature extraction and logit prediction, such that we adapt the features for the input $\vq$ and the exemplar memory $V$: $\tilde{\vq} = h(\vq, V)$.
The output $\tilde{\vq}$ is then passed to a classifier to generate logits: $\vz = g(\tilde{\vq})$.
The \MTN model is shown in Figure~\ref{fig:overview}.

The transformer architecture \cite{vaswani2017attention} is a natural candidate for $h$. 
Transformers map a (possibly ordered) set of input features to a set of output features, where each output feature is dependent on all the input features through self-attention. 
This fits the modeling requirement for incorporating the exemplar memory set $V$ into the prediction process.
We want the adapted features to be able to attend to $\vq$ and also to the exemplar memory features $V$.

The exemplar memory set $V$ can grow to be of the order of tens of thousands of exemplars. 
Therefore, we only provide a subset of $V$ as context for a given input $\vq$. 
To choose a subset of $V$ for given input $\vq$, we first find the $k$-nearest-neighbors (kNNs) $\vq$ from the memory $V$.
Thus the transformer is tasked with performing a local attention of the input feature vector $\vq$, dependent on its local neighborhood in the feature space.

%% file: fig/overview.tex
\begin{figure}[t]
\begin{center}
\includegraphics[width=0.90\linewidth]{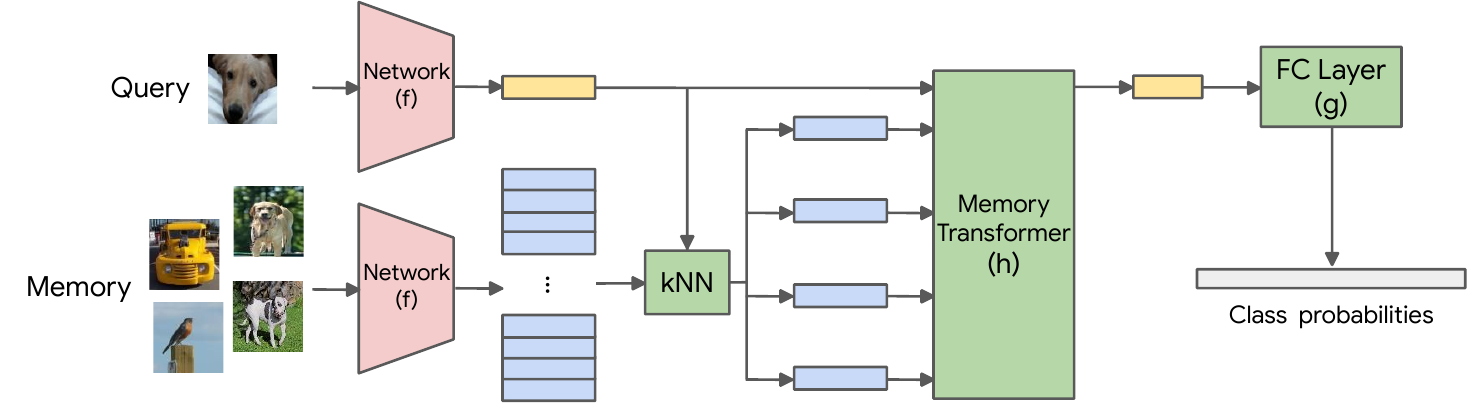}
\end{center}
\caption{
\textbf{Overview of the MTN model.} A query feature vector is provided as input to the \MTN, along with the $k$ nearest neighbors from a set of exemplar memory feature vectors. \MTN adapts the input query feature by learning to model the relationship between the query vector and its nearest neighbors. We then pass the adapted feature to a linear head to calculate a set class scores (or logits).
}
\label{fig:overview}
\end{figure}

%% file: sec/5_results_bmvc.tex
\section{Experiments}
In this section, we first describe the experimental setup and implementation details.
Then we compare our method to existing approaches for class-incremental learning.
We also present detailed ablation studies on \MTN showing the impact of different design choices.

\subsection{Experimental Setup}
\head{Datasets. }
We assess \MTN for class-based incremental learning separately on two datasets: ImageNet-1K~\cite{DSLLF09} and Google Landmarks-1K~\cite{ahn2020ssil}.
Following standard practices~\cite{rebuffi2017icarl}, we use the \emph{average incremental accuracy} as the evaluation metric, which is the top-$1$ accuracy averaged over $T$ tasks.
We now describe both datasets in more detail.

ImageNet-1k contains $C=1000$ classes, $1.2$ million training images, and  $50k$ validation images. 
We follow the work of Hou~\etal~\cite{hou2019rebalance} and create the incremental tasks using the following approach.
The first task contains half of the classes, \ie $C/2 = 500$ classes.
We then divide rest of the classes into tasks of $L=50$, $L=100$ or $L=250$ classes.

Google Landmarks-1K is a subset of $C=1000$ classes from Google Landmarks v2~\cite{weyand2020google}, which originally contains $203,094$ classes.
We follow the work of Ahn~\etal~\cite{ahn2020ssil}, and sample $1000$ classes in the order of largest sample of images per class.
Similarly to ImageNet-1K, the first task, contains $500$ classes, and the other tasks are divided into $L=50$, $L=100$ or $L=250$ classes.

\head{Feature representations. }
In addition to the features learned from scratch with ResNet-18 backbone, we also use \emph{fixed} features rather than back-propagating through the entire backbone, similarly to ~\cite{kemker2018fearnet,xiang2019adversarial}. However, we argue that using features that were pretrained for classification in a supervised way~\cite{kemker2018fearnet,xiang2019adversarial} might confound our results since the backbone would have knowledge about all tasks at all times.
For this reason, we choose to use representations obtained from no (or weak) supervision~\cite{chen2020simple,radford2021clip}.
Actually, these representations have been shown to be generic and excellent ``off-the-shelf'' descriptors, performing well on a large variety of tasks and datasets without any fine-tuning~\cite{caron2021emerging,zhai2021lit}.

In practice, we use CLIP~\cite{radford2021clip} (ViT-B/32) features of $d=512$, and SimCLR~\cite{chen2020simple} (ResNet-50) features of $d=2048$ as \emph{fixed} features.
For \emph{learned} features, we use a ResNet-18 ($d=512$) backbone learned with SS-IL~\cite{ahn2020ssil} and updated at each task. 
We learn the classifier separately after the feature extractor is learned and its weights are frozen.

\begin{table}[t]
  \input{tab/incremental}
  \caption{\textbf{Comparison with existing works.}
  We report the average incremental accuracy of \MTN and other methods.
  We vary the number of classes in each task after the base task ($L=50, 100, 250$).
  Reported numbers use different architectures: ResNet-18 (``R18''), ResNet-50 pre-trained with SimCLR (``$\text{R50}_\text{(SimCLR)}$'') or ViT-Base/32 pretrained with CLIP (``$\text{ViT-B}_\text{(CLIP)}$'').
  \faLock~means that the features are kept fixed when learning the classifier.
  We specify when the memory is used during training as a source of additional data (``Train.'') and explicitly during the prediction process (``Pred'').
  $\dagger$: other methods run by us.
  \label{tab:incremental}}
\end{table}

\head{Implementation details. }
We train \MTN with SGD optimizer, a learning rate of $0.1$, weight decay of $0.0001$ and momentum of $0.9$. 
We train the model for $10$ epochs per task.
We use a batch size of $128$, but also add $32$ samples from the previous tasks to each batch~\cite{ahn2020ssil}. 
We use the Ringbuffer method~\cite{chaudhry2019continual} to select the exemplars and construct the memory.
There are $M$ vectors in the memory at any given point.
As new classes become available, some of the exemplars from the existing classes are removed to maintain a memory of size $M$.
We typically set the memory size to $M=20k$ and select top $k=10$ nearest neighbors of each query as the input to the transformer.
Our default setting is a multi-head self-attention transformer with $4$ layers, $4$ attention heads, and the output channel dimension of $128$.
We will release code and checkpoints to reproduce our results.

\begin{figure}[t]
\input{fig/task_acc}
\caption{\textbf{Task accuracy.} Top-1 accuracy after each task is shown for \MTN and baselines in ImageNet-1k dataset with SimCLR features for different number of classes per task ($L$).
\label{fig:taskAcc}
}
\vspace{-2pt}
\end{figure}
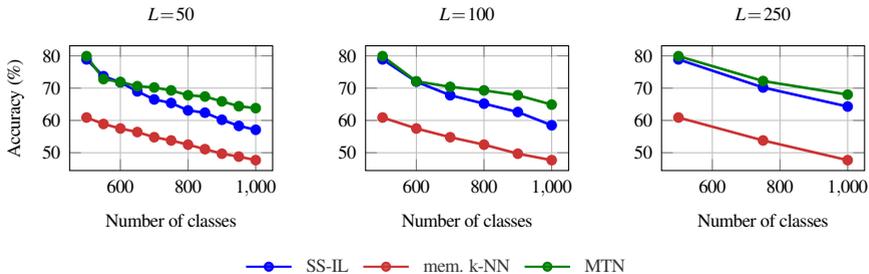

\subsection{Comparison with Existing Works}

We demonstrate the effectiveness of \MTN on ImageNet-1k and Landmarks-1k, two standard incremental learning benchmarks in Table~\ref{tab:incremental}.
For fair comparisons and to avoid confounding factors, we re-implement different methods~\cite{wu2019large,rebuffi2017icarl,hou2019rebalance,ahn2020ssil,mittal2021essentials} in a similar setup as \MTN, i.e. with fixed pre-trained features from SimCLR and CLIP.

We propose and report a simple baseline, ``mem. k-NN'', that performs classification based on nearest neighbor search within the memory of exemplars.
We set $k=10$ for ``mem. k-NN'', as it gives the highest performance.
We see that our \MTN approach performs much better than the ``mem. k-NN'' baseline, $+21\%$ with SimCLR features (\faLock~-$\text{R50}_\text{(SimCLR)}$), demonstrating the effectiveness of learning \emph{how} to aggregate information from the memory for prediction.

Overall, we observe in Table~\ref{tab:incremental} that \MTN outperforms other existing methods by significant margins when trained on top of fixed representations and sets a new state-of-the-art for class-incremental learning on ImageNet-1k.
The comparison with SS-IL is particularly interesting as we re-use the framework and loss proposed by SS-IL to train \MTN, as described in Section~\ref{sec:background}. 
Concretely, the only difference between SS-IL and ours is the architecture of the classifier;
SS-IL uses a linear layer while we use a transformer conditioned on the elements of the memory bank.
We observe in Table~\ref{tab:incremental} that \MTN improves over SS-IL, showing the effectiveness of our prediction process based on memory transformer.

Consistently over all approaches, we see that using SimCLR features perform better on ImageNet-1k than Landmarks-1k.
We hypothesize that this is due to the fact than ResNet-50 was pre-trained on ImageNet-1k dataset (without labels) and hence the representations are better suited for classification on this dataset whereas there is a domain discrepancy when applied to Landmarks-1k.

We also show that \MTN can easily be applied to features learned from scratch. For ResNet-18 experiments, we train the feature extractor from scratch using SS-IL, and freeze the feature extractor at the end of each task. We then learn \MTN as the classifier.
Table~\ref{tab:incremental} shows that \MTN trained on top of ResNet-18 SS-IL features (\faLock~-$\text{R18}_\text{(SS-IL)}$) achieves up to $+4\%$ relative improvement compared to SS-IL without \MTN.
\MTN applied on top of SS-IL does not outperform more recent baselines on ImageNet-1k, but we note that \MTN architecture is not specific to SS-IL, can be applied to any other number of approaches, such as PODNet~\cite{douillard2020podnet} or CCIL~\cite{mittal2021essentials}. 
On the other hand, \MTN applied on SS-IL features achieves a new state-of-the-art for Landmarks-1k dataset.

In Figure~\ref{fig:taskAcc}, we show the accuracy on all seen classes after each task.
We report results for MTN and two baselines: ``mem. k-NN'' (that uses elements of the memory for prediction based on nearest neighbors) and SS-IL~\cite{ahn2020ssil} which is trained similarly as ours but with a linear classifier instead of the MTN.
We see that the accuracy for the SS-IL baseline starts at a similar level to \MTN, but decreases at a faster pace as tasks are seen, which shows that it severely suffers from catastrophic forgetting.
On the other hand, the accuracy of \MTN does not drop quickly, resulting in a higher overall incremental accuracy.

\subsection{Ablation studies}
In this section, we evaluate different design choices of \MTN like the number of nearest neighbors to select from the memory, details of the transformer architecture or the use of positional encoding.
We also propose a qualitative evaluation of what is learned by \MTN.
All experiments are conducted with the SimCLR features (\faLock~-$\text{R50}_\text{(SimCLR)}$).

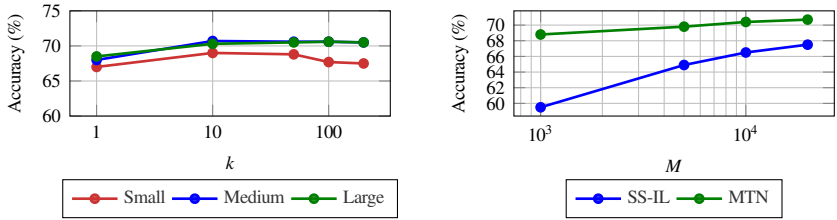
\begin{figure}[t]
\input{fig/fig_k_size}
\caption{\textbf{Left: Impact of $k$ and the architecture.} The average incremental accuracy for \MTN is reported for different $k$ and transformer architectures with $L=100$ classes per task on ImageNet-1k. \textbf{Right: Impact of the memory size $M$.} Average incremental accuracy is shown for \MTN and SS-IL for $L=100$ classes per task on ImageNet-1k.
\label{fig:ksize}
}
\end{figure}

\head{The number of kNN. }
We vary $k$ the number of nearest neighbors to input to \MTN.
More precisely, the hyperparameter $k$ is used to obtain the k-NN list $\nn{}(\vq; V)$ for a given query vector $\vq$.
As mentioned above, choosing $k$ nearest neighbors from the memory is essential and keeps \MTN efficient, as it becomes impractical to input the entire memory of $M=20k$ vectors to the transformer. 
Figure~\ref{fig:ksize} (Left) shows the impact of different $k$ for \MTN.
We observe that the accuracy of \MTN remains stable for different $k$ as long as it is ``large enough'' (more than $10$).
We set $k=10$ for all of our experiments.

\head{Memory size. }
We vary the size of the memory $M$, and report the average incremental accuracy for \MTN and SS-IL in Figure~\ref{fig:ksize} (Right). 
We observe that \MTN is more robust to smaller $M$, where as the accuracy for SS-IL rapidly decreases for smaller $M$.
This shows that \MTN is an excellent candidate for efficient incremental learning systems requiring very small memory footprint.

\head{Transformer size.}
We examine the effect of the different transformer architectures in Figure~\ref{fig:ksize} (Left). 
We compare across 3 architectures.
The ``small'' architecture has $2$ layers, $1$ attention head, and the output channel dimension of $64$. 
The ``medium'' architecture has $4$ layers and attention heads, and the output channel dimension of $128$.  
The ``large'' architecture has $12$ layers and attention heads, and the output channel dimension of $768$.   
The ``small'' is less robust to the different values of $k$.
There is no benefit going from ``medium'' to ``large'', as they perform similarly.
We use the ``medium'' for all of our experiments.

\begin{figure*}[t]
\begin{center}
\includegraphics[width=0.85\linewidth]{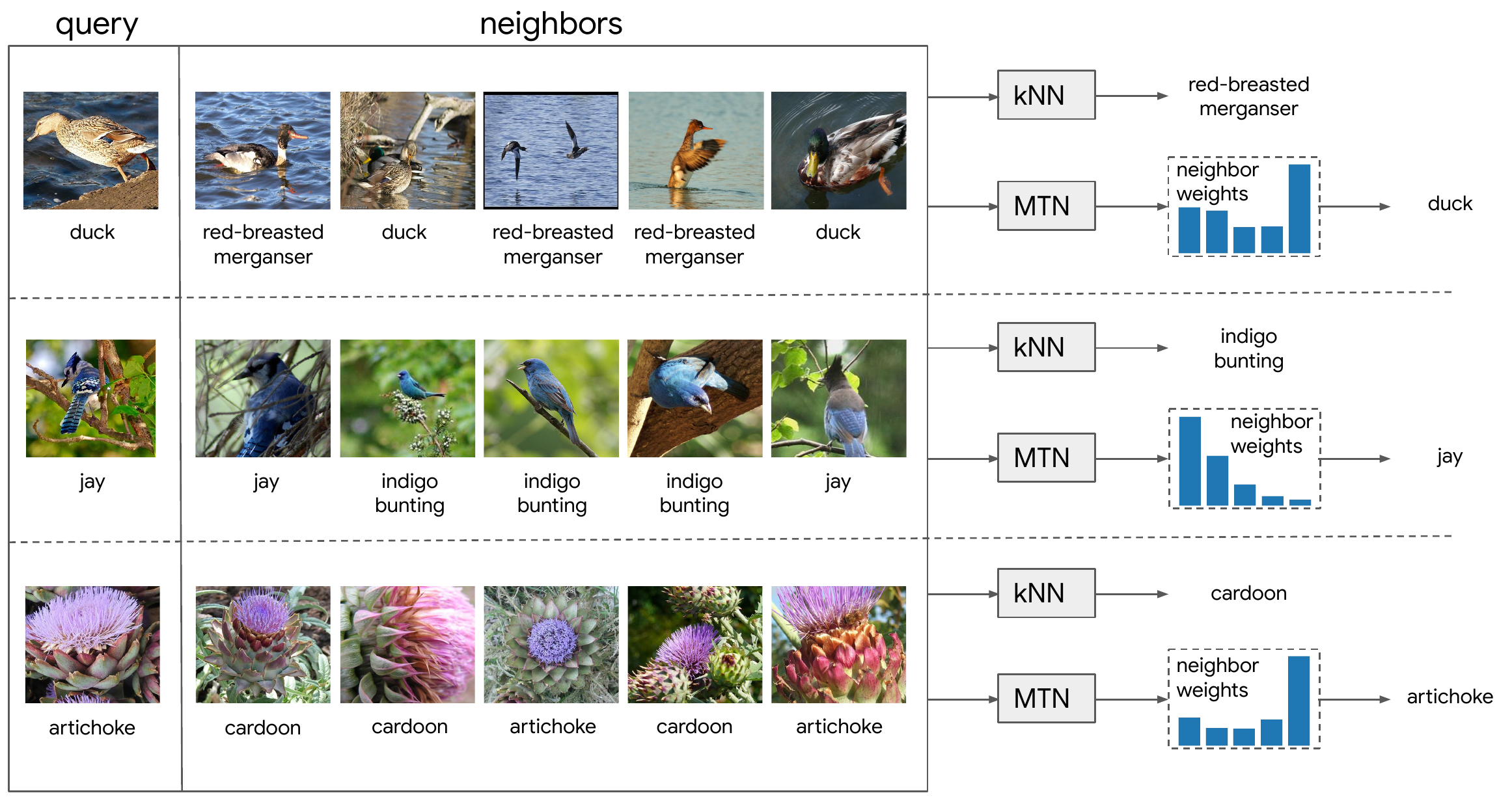}
\end{center}
\caption{
\textbf{Qualitative examples.} A visualization of examples from ImageNet-1K which kNN incorrectly classifies but \MTN correctly classifies. We first show the kNN returned by the initial search. Note that the neighbors are ordered, with closest to the query on the left. 
We adapt the query and kNN features with \MTN and show the similarity of the adapted features in the blue histogram, where each bar corresponds to a neighbor.
}
\label{fig:knnvis}
\vspace{-10pt}
\end{figure*}

\head{Qualitative results.}
We present a qualitative visualization to inspect the behavior of \MTN and get an intuition of how it learns to make more accurate predictions.
In Figure~\ref{fig:knnvis}, we look at how the similarity with different exemplars evolves after being processed by \MTN.
For each query image, we show its five nearest neighbors within the exemplars in the memory, i.e. before \MTN processing, and report the prediction based on the vote of these neighbors.
This is similar to baseline ``mem. k-NN'' in Table~\ref{tab:incremental}.
Then, in the ``MTN'' rows of Figure~\ref{fig:knnvis}, we show the similarity of these exemplars with the query as processed by \MTN.
Concretely we show in the histogram, the similarity between the \MTN-adapted query vector $\vq$, and the \MTN-adapted k-NN features $\tilde{V}_{\nn{M}(\vq)}$.
We observe in Figure~\ref{fig:knnvis} how the features adapted by \MTN have better similarity.

For sets of neighbors with the incorrect label as the majority, \MTN is able to successfully up-weight neighbors that do have the correct label.
We find it interesting to note that, even for images which appear at a high-level to be similar, \MTN assigns most of the weight to one or two images.
This suggests that \MTN is able to successfully discriminate between confounding examples.

%% file: tab/incremental.tex
\scriptsize
\setlength{\tabcolsep}{3pt}
\begin{center}
\begin{tabular}{ll@{\msp}cc@{\msp}c@{\msp}ccc@{\msp}ccc}
\toprule
Method	&& Backbone	& \multicolumn{2}{c}{Memory usage:} & \multicolumn{3}{@{\lsp}c@{\lsp}}{ImageNet-1k}&	 \multicolumn{3}{@{\msp}c@{\msp}}{Landmarks-1K} 	\\
								&&	 & 	Train. & Pred.	& $50$		& $100$		& $250$		& $50$		& $100$		& $250$		\\
\midrule

PODNet~\cite{douillard2020podnet}	&& $\text{R18}_\text{(Scratch)}$ &\checkmark 	& 					& 64.1		& 67.0		& -		& -		& -		& -		\\
CCIL~\cite{mittal2021essentials}$\dagger$	&& $\text{R18}_\text{(Scratch)}$ &\checkmark 	& 					& \textbf{65.8}		& \textbf{67.6}		& \textbf{69.1}	& 40.1		& 49.7		& 55.4 	\\
SS-IL~\cite{ahn2020ssil}$\dagger$	    	&&  $\text{R18}_\text{(Scratch)}$  & \checkmark 	&				& 57.0		& 62.5		& 67.1	& 54.9	& 60.0	& 64.7	\\
\rowcolor{Light}
\MTN	    	&&  \faLock~$\text{R18}_\text{(SS-IL)}$  & \checkmark 	& \checkmark	& 60.8		& 64.8		& 68.4 	&  \textbf{59.4}	& \textbf{63.1}  & \textbf{66.5}  \\
\midrule 
mem. k-NN$\dagger$						&& \faLock~-$\text{R50}_\text{(SimCLR)}$   	&  & \checkmark	& 53.8		& 53.9		& 54.2		& 30.0		& 30.1		& 30.2		\\
BIC~\cite{wu2019large}$\dagger$			&& \faLock~-$\text{R50}_\text{(SimCLR)}$  	& \checkmark &	& 34.3		& 43.9		& 56.9		& 17.4		& 26.5		& 38.0		\\
iCARL~\cite{rebuffi2017icarl}$\dagger$	&& \faLock~-$\text{R50}_\text{(SimCLR)}$   	& \checkmark & \checkmark	& 56.5		& 57.0		& 58.4		& 27.3		& 27.7		& 28.5		\\
LUCIR~\cite{hou2019rebalance}$\dagger$ 	&& \faLock~-$\text{R50}_\text{(SimCLR)}$   	& \checkmark &	& 60.1		& 63.0		& 68.7		& 48.4		& 50.9		& 55.3		\\
SS-IL~\cite{ahn2020ssil}$\dagger$		&& \faLock~-$\text{R50}_\text{(SimCLR)}$   	& \checkmark &	& 66.1		& 67.5		& 71.2		& 46.2		& 46.6		& 47.9		\\
\rowcolor{Light}
\MTN									&& \faLock~-$\text{R50}_\text{(SimCLR)}$   & \checkmark &	\checkmark & \textbf{69.5}		& \textbf{70.7}		& \textbf{73.3}		& \textbf{51.0}		& \textbf{52.3}		& \textbf{55.4}	\\
\midrule 
mem. k-NN$\dagger$								&& \faLock~-$\text{ViT-B}_\text{(CLIP)}$   & 	  & \checkmark	& 47.7		& 47.7		& 48.1		& 35.8		& 35.9		& 36.1		\\
BIC~\cite{wu2019large}$\dagger$			&& \faLock~-$\text{ViT-B}_\text{(CLIP)}$   & \checkmark &			& 32.9		& 43.2		& 57.5		& 25.3		& 35.6		& 47.2		\\
iCARL~\cite{rebuffi2017icarl}$\dagger$	&& \faLock~-$\text{ViT-B}_\text{(CLIP)}$   & \checkmark & \checkmark			& 52.3		& 53.2		& 54.7		& 36.5		& 36.9		& 37.8		\\
LUCIR~\cite{hou2019rebalance}$\dagger$ 	&& \faLock~-$\text{ViT-B}_\text{(CLIP)}$   & \checkmark &			& 51.6		& 55.7		& 63.3		& 46.2		& 48.8		& 53.8		\\
SS-IL~\cite{ahn2020ssil}$\dagger$		&& \faLock~-$\text{ViT-B}_\text{(CLIP)}$   & \checkmark &			& 63.9		& 65.7		& 69.7		& 52.4		& 52.8		& 54.4		\\
\rowcolor{Light}
\MTN							&& \faLock~-$\text{ViT-B}_\text{(CLIP)}$   & 	 \checkmark & \checkmark	& \textbf{67.4}		& \textbf{68.9}		& \textbf{71.2}		& \textbf{54.1}		& \textbf{55.5}		& \textbf{57.6}	\\

\bottomrule
\end{tabular}
\end{center}

%% file: fig/task_acc.tex
\newenvironment{customlegend}[1][]{%
    \begingroup
    \csname pgfplots@init@cleared@structures\endcsname
    \pgfplotsset{#1}%
}{%
    \csname pgfplots@createlegend\endcsname
    \endgroup
}%

\def\addlegendimage{\csname pgfplots@addlegendimage\endcsname}

\centering
\input{fig/data/sample}
\begin{tabular}{ccc}
{

\begin{tikzpicture}
    \tikzstyle{every node}=[font=\scriptsize]
\begin{axis}[%
  width=0.33\textwidth,
  height=0.25\textwidth,
  xlabel={Number of classes},
    grid=both,
  ylabel= {Accuracy (\%)},
  xminorticks=false,
  title={$L=50$},
    legend cell align={left},
    legend pos=outer north east,
    legend style={at={(0.5,-0.5)},anchor=north,legend columns=-1, cells={anchor=west}, font ={\footnotesize}, fill opacity=0.8, row sep=-2.5pt},
]

  \addplot[color=red,     solid, mark=*,  mark size=1.5, line width=1.0] table[x=C, y expr={\thisrow{knn}}] \taskAccFifty;
  \addplot[color=blue,     solid, mark=*,  mark size=1.5, line width=1.0] table[x=C, y expr={\thisrow{linear}}] \taskAccFifty;
  \addplot[color=dark_green,     solid, mark=*,  mark size=1.5, line width=1.0] table[x=C, y expr={\thisrow{mtn}}] \taskAccFifty;

\end{axis}
\end{tikzpicture}

}

&

{

\begin{tikzpicture}
    \tikzstyle{every node}=[font=\scriptsize]
\begin{axis}[%
  width=0.33\textwidth,
  height=0.25\textwidth,
  xlabel={Number of classes},
    grid=both,
  xminorticks=false,
  title={$L=100$},
    legend cell align={left},
    legend pos=outer north east,
    legend style={at={(0.5,-0.5)},anchor=north,legend columns=-1, cells={anchor=west}, font ={\footnotesize}, fill opacity=0.8, row sep=-2.5pt},
]

  \addplot[color=red,     solid, mark=*,  mark size=1.5, line width=1.0] table[x=C, y expr={\thisrow{knn}}] \taskAccHundred;
  \addplot[color=blue,     solid, mark=*,  mark size=1.5, line width=1.0] table[x=C, y expr={\thisrow{linear}}] \taskAccHundred;
  \addplot[color=dark_green,     solid, mark=*,  mark size=1.5, line width=1.0] table[x=C, y expr={\thisrow{mtn}}] \taskAccHundred;

\end{axis}
\end{tikzpicture}

}

&

{

\begin{tikzpicture}
    \tikzstyle{every node}=[font=\scriptsize]
\begin{axis}[%
  width=0.33\textwidth,
  height=0.25\textwidth,
  xlabel={Number of classes},
    grid=both,
  xminorticks=false,
  title={$L=250$},
    legend cell align={left},
    legend pos=outer north east,
    legend style={at={(0.5,-0.5)},anchor=north,legend columns=-1, cells={anchor=west}, font ={\footnotesize}, fill opacity=0.8, row sep=-2.5pt},
]

  \addplot[color=red,     solid, mark=*,  mark size=1.5, line width=1.0] table[x=C, y expr={\thisrow{knn}}] \taskAccTwoHundredFifty;
  \addplot[color=blue,     solid, mark=*,  mark size=1.5, line width=1.0] table[x=C, y expr={\thisrow{linear}}] \taskAccTwoHundredFifty;
  \addplot[color=dark_green,     solid, mark=*,  mark size=1.5, line width=1.0] table[x=C, y expr={\thisrow{mtn}}] \taskAccTwoHundredFifty;

\end{axis}
\end{tikzpicture}

}

\end{tabular}

\centering
\begin{tikzpicture}
\scriptsize
\begin{customlegend}[legend columns=4,legend style={align=left,draw=none,column sep=1ex},
        legend entries={SS-IL,
                        mem. k-NN,
                        \MTN,
                        }]
        \addlegendimage{color=blue, solid, mark=*, mark size=1.5}
        \addlegendimage{color=red, solid, mark=*, mark size=1.5}
        \addlegendimage{color=dark_green, solid, mark=*, mark size=1.5}
        \end{customlegend}
\end{tikzpicture}

%% file: fig/fig_k_size.tex


\input{fig/data/sample}
\begin{tabular}{cc}
{

\begin{tikzpicture}
    \tikzstyle{every node}=[font=\scriptsize]
\begin{axis}[%
  width=0.45\textwidth,
  height=0.23\textwidth,
  xlabel={$k$},
    xmode=log,
    ymin=60,
    ymax=75,
    grid=both,
    xtick={1,10,100,500},
    xticklabels={1,10,100,500},
  ylabel= {Accuracy (\%)},
  xminorticks=false,
    legend cell align={left},
    legend pos=outer north east,
    legend style={at={(0.5,-0.6)},anchor=north,legend columns=-1, cells={anchor=west}, font ={\footnotesize}, fill opacity=0.8, row sep=-2.5pt},
]

  \addplot[color=red,     solid, mark=*,  mark size=1.5, line width=1.0] table[x=k, y expr={\thisrow{small}}] \knnSizeComp;\leg{Small};
  \addplot[color=blue,     solid, mark=*,  mark size=1.5, line width=1.0] table[x=k, y expr={\thisrow{medium}}] \knnSizeComp;\leg{Medium};
  \addplot[color=dark_green,     solid, mark=*,  mark size=1.5, line width=1.0] table[x=k, y expr={\thisrow{large}}] \knnSizeComp;\leg{Large};

\end{axis}
\end{tikzpicture}

}

&

{

\begin{tikzpicture}
    \tikzstyle{every node}=[font=\scriptsize]
\begin{axis}[%
  width=0.45\textwidth,
  height=0.23\textwidth,
  xlabel={$M$},
    xmode=log,
    grid=both,
  ylabel= {Accuracy (\%)},
  xminorticks=false,
    legend cell align={left},
    legend pos=outer north east,
    legend style={at={(0.5,-0.6)},anchor=north,legend columns=-1, cells={anchor=west}, font ={\footnotesize}, fill opacity=0.8, row sep=-2.5pt},
]

  \addplot[color=blue,     solid, mark=*,  mark size=1.5, line width=1.0] table[x=M, y expr={\thisrow{hundred}}] \linearMemoryComp;\leg{SS-IL};
  \addplot[color=dark_green,     solid, mark=*,  mark size=1.5, line width=1.0] table[x=M, y expr={\thisrow{hundred}}] \memoryComp;\leg{\MTN};

\end{axis}
\end{tikzpicture}

}

\end{tabular}


%% file: sec/6_conclusions_bmvc.tex
\section{Conclusion}

We introduce \MTN, a method to improve exemplar-based class-incremental learning performance by utilizing the exemplars directly in the prediction process.
We demonstrate that \MTN allows to consistently improve class-incremental learning performance in various benchmarks.
We show experimentally that \MTN with generic pretrained features is competitive with state-of-the-art methods.
In future work, we would like to evaluate \MTN with unlocked features and train the full architecture end-to-end in an incremental manner.